\documentclass{article}
\usepackage{spconf,amsmath,amssymb,graphicx,hyperref}
\usepackage[most]{tcolorbox}
\usepackage{booktabs}
\usepackage{multirow}
\usepackage[table]{xcolor}
\usepackage{colortbl}
\DeclareUnicodeCharacter{266B}{}

\title{Logic-ORiented Retriever Enhancement via Contrastive Learning}
\name{Wenxuan Zhang$^{1}$, Yuan-Hao Jiang$^{1}$, Changyong Qi$^{1}$, Rui Jia$^{1}$, Yonghe Wu$^{2}$\thanks{Corresponding author: yhwu@deit.ecnu.edu.cn}}
\address{$^{1}$Shanghai Institute of Artificial Intelligence for Education, East China Normal University, China\\
$^{2}$Education Technology, East China Normal University, China}
\begin{document}
%
\maketitle
\begin{abstract}
Large language models (LLMs) struggle in knowledge-intensive tasks, as retrievers often overfit to surface similarity and fail on queries involving complex logical relations. The capacity for logical analysis is inherent in model representations but remains underutilized in standard training.  LORE (Logic ORiented Retriever
Enhancement) introduces fine-grained contrastive learning to activate this latent capacity, guiding embeddings toward evidence aligned with logical structure rather than shallow similarity. LORE requires no external supervision, resources, or pre-retrieval analysis, remains index-compatible, and consistently improves retrieval utility and downstream generation while maintaining efficiency. The datasets and code are publicly available at \url{https://github.com/mazehart/Lore-RAG}.
\end{abstract}
\section{Introduction}

Large language models (LLMs) have achieved significant progress across a wide range of natural language processing tasks~\cite{DBLP:conf/nips/BrownMRSKDNSSAA20, arslan2024survey}, yet they remain limited in knowledge-intensive scenarios, often exhibiting factual hallucinations, stale knowledge, and poor traceability. These limitations reduce reliability and make it difficult to deploy LLMs in high-stakes applications where correctness and verifiability are critical \cite{huang2025survey, DBLP:conf/emnlp/LiCZNW23}. Therefore, retrieval-augmented generation (RAG) has emerged as a practical remedy~\cite{lewis2020retrieval}. By following a "retrieve-then-generate" paradigm, RAG retrieves external knowledge sources through embedding models via vector matching and then passes them to large models, thereby mitigating hallucination problems and enhancing interpretability and source traceability \cite{jiang2023structgpt,sarthi2024raptor}.

\begin{figure}[htbp]
    \centering
    \includegraphics[width=0.47\textwidth]{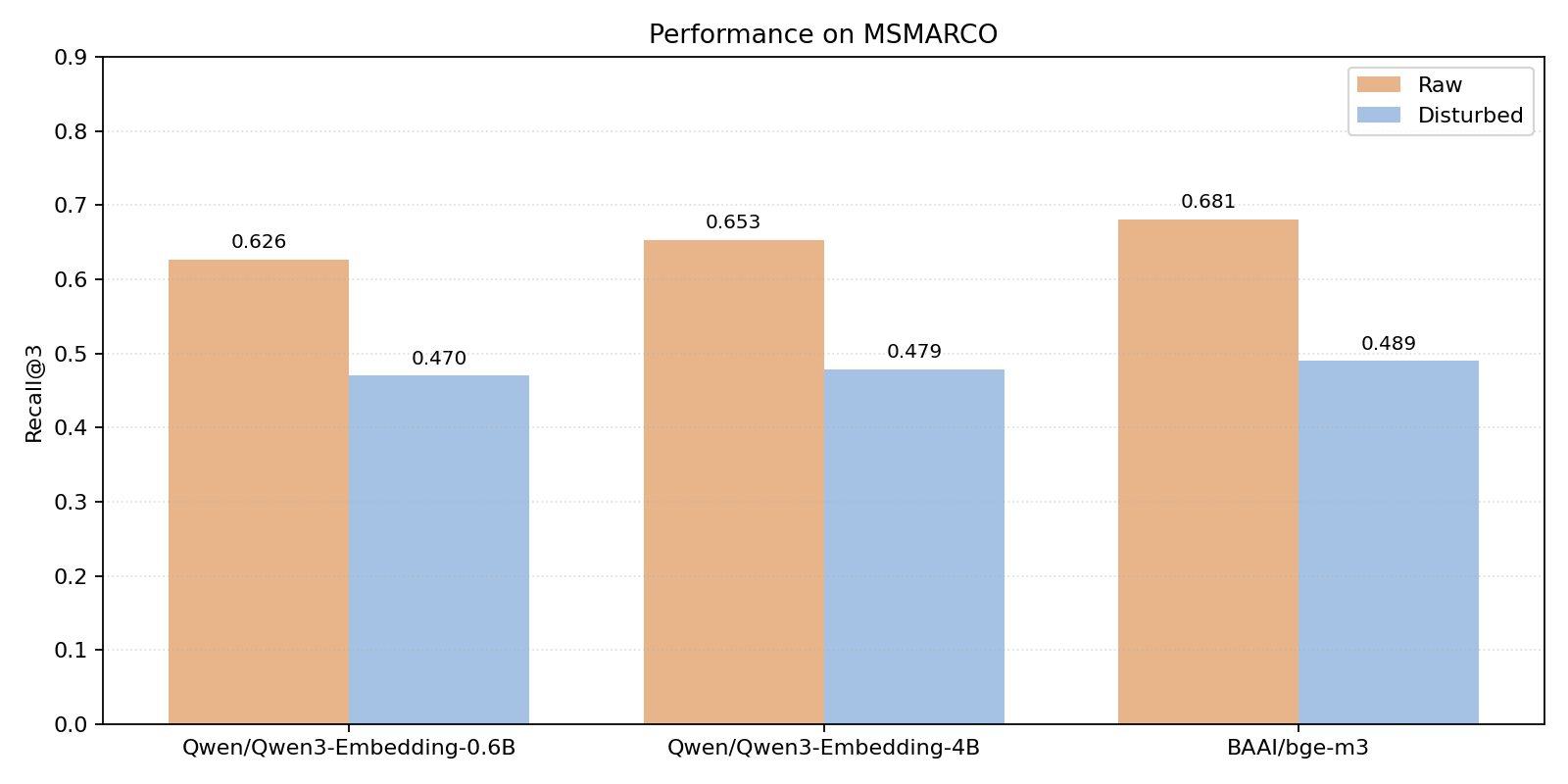}
    \caption{\textbf{Impact of complex logical expressions on embedding model performance.} Results on MSMARCO dataset show significant performance degradation across different embedding models when queries are disturbed with complex logical expressions, highlighting the vulnerability of current embedding approaches to complex logical structures.}
\end{figure}

In practice, a persistent gap exists: while embedding models perform well on direct and straightforward queries, their effectiveness significantly degrades when handling queries disturbed with complex logical expressions. As shown in the figure, existing open-source embedding models, even with large parameter counts, cannot avoid the difficulties brought by queries disturbed with complex logical expressions. This is because their similarity-driven nature causes them to often favor surface-level lexical overlap rather than faithfully representing and satisfying the complex logical requirements of queries, leading to fragile evidence sets that fail to meet query requirements \cite{caspari2024beyond}.

A natural response is to analyze or decompose queries using powerful LLMs before retrieval \cite{agrawal2024mindful, gan2024similarity}. However, this "LLM-in-the-loop" strategy introduces non-trivial costs and latency, and relies on knowledge graphs and other structures constructed for datasets. This paper takes a different path: rather than adding more LLM computation, we pose a question—do existing open-source embedding models already contain untapped capacity to represent complex linguistic structures? Can targeted contrastive learning activate this capacity, enabling retrievers to directly encode complex logical structures and salience rather than relying solely on surface similarity?
\begin{figure*}[htbp]
    \centering
    \includegraphics[width=\textwidth]{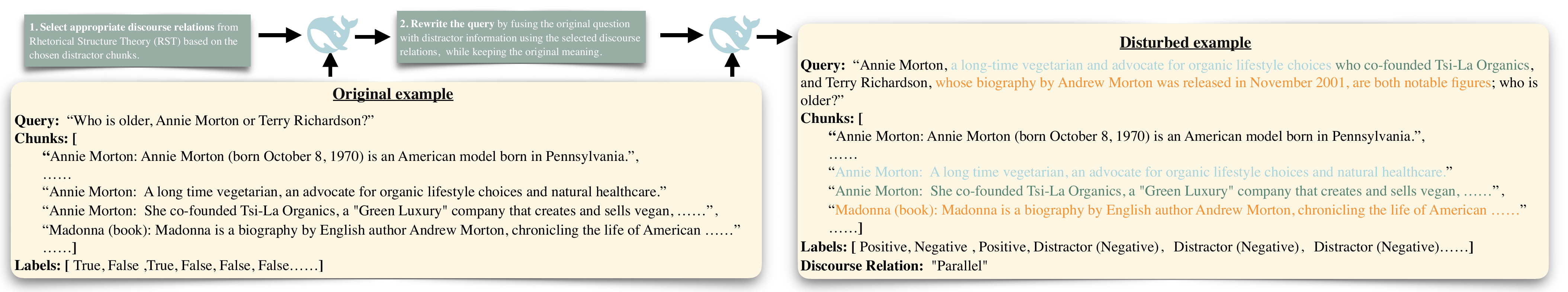}
    \caption{An illustration of \textbf{Query Rewriting}. The left side shows an original example with query, chunks, and labels, while the right side presents the corresponding modified example with enriched query, discourse relation, and distractor annotations.}
    \label{fig:Query Rewriting}
\end{figure*}

This paper presents LORE, an embedding enhancement method through contrastive learning for improving retrievers. Based on existing datasets, LORE does not invoke large models to assist embedding after queries and before retrieval, but rather retrains the embedding model using fine-grained contrastive knowledge annotated by large models after pretraining, emphasizing signals related to complex logical expressions.

The contributions of this paper are three-fold:
\begin{itemize}
    \item This paper proposes LORE, an embedding model enhancement paradigm that integrates large model knowledge in the post-pretrain stage, providing a simple contrastive fine-tuning method.
    \item This paper constructs an open-source fine-grained contrastive learning dataset based on existing open-source.
    \item This paper validates the consistent gains of LORE on multiple tasks featuring queries disturbed with complex logical expressions, with improvements on unlabeled tasks particularly demonstrating the effectiveness of the proposed method.
\end{itemize}

\section{Related Work}

\noindent\textbf{Beyond similarity.} Recent methods have begun to address the limitations caused by RAG's reliance on similarity. METRAG attempts to use changes in LLM answer accuracy on downstream tasks as supervision signals to fine-tune the retrieval embedding model, training it to identify retrieval content that helps improve LLM answer accuracy on downstream tasks \cite{gan2024similarity}. Meanwhile, Mindful-RAG combines LLM's internal knowledge with external knowledge graphs, first analyzing query intent through LLMs, then integrating knowledge graphs to introduce context alignment and result verification in the retrieval process, thereby reducing errors and ambiguity in complex reasoning \cite{agrawal2024mindful}.

\noindent\textbf{Limitations.} These methods improve utility through task feedback, intent modeling, or external knowledge structures, but often introduce additional overhead and rely on external supervision or resources, hindering transferability across different tasks. While they leverage large model knowledge to enhance embedding model performance, the fundamental capabilities of embedding models are not improved in this process. Therefore, a key question is: how to equip retrievers with understanding of queries with complex logical expressions, enabling them to directly encode queries and prioritize evidence that conforms to the query's logical structure, while maintaining efficiency and avoiding dependence on pre-retrieval LLM analysis, external supervision, and external knowledge structures \cite{caspari2024beyond}.

\noindent\textbf{Positioning.} This paper proposes LORE, a contrastive post-training method to activate the latent capacity of open-source embedding models in representing queries with complex logical expressions, enabling retrievers to directly encode logical structure rather than relying on surface similarity, thereby improving retrieval performance at the fundamental capability level of embeddings.

\section{Method}

\subsection{Dataset Construction}

\noindent\textbf{Fine-grained Chunk Categories}
The construction starts from a base corpus with query $q$, candidate chunks $\{c_k\}$, and boolean labels $y_k\in\{\text{True},\text{False}\}$. For each $q$, a subset of False-labeled chunks is sampled as distractors, each selected distractor is assigned its original chunk index as a source identifier, and an LLM is prompted to rewrite $q$ into a more natural query that embeds some distractor content while preserving the original meaning. The distractor usage is then mapped back to chunk indices and three tiers are constructed for each $q$ and candidate set $\mathcal{C}(q)$:

- \textit{Positive (P)}: chunks with $y_k=\text{True}$ that are sufficient to answer the query.

- \textit{Distractor (N1)}: False-labeled chunks that were used by the LLM in the rewritten question as distractors, appearing relevant to the question but providing no help in answering it.

- \textit{Negative (N2)}: other False-labeled chunks not used in the rewrite.

The three tiers are denoted as P, N1, and N2. When only a portion of the dataset is rewritten, the remaining samples retain their original queries and boolean labels, which are mapped to these tiers for consistency.

\noindent\textbf{Query Rewriting}
The dataset construction follows a two-step process as illustrated in Figure~\ref{fig:Query Rewriting}. First, appropriate discourse relations from Rhetorical Structure Theory (RST) \cite{mann1988rst} are selected based on the chosen chunks. Eight common relations—Sequential, Transitional, Supplementary, Contrastive, Causal, Parallel, Hypothetical, and Explanatory—serve as guidelines. Second, the query is rewritten by fusing the original question with distractor information using the selected discourse relations, resulting in a modified example with enriched query and discourse relation annotations. The constructed training dataset, CoEnTrain, is released.

\subsection{Contrastive Learning}

\noindent\textbf{Contrastive loss}
The framework employs two separate encoders: a query encoder $\mathcal{M}_q$ and a document encoder $\mathcal{M}_d$. Queries and chunks are embedded into a shared vector space using the respective encoders. Let $\mathbf{h}_q = \mathcal{M}_q(q)$ and $\mathbf{h}_k = \mathcal{M}_d(c_k)$ denote the normalized embeddings, and $s_k$ the cosine similarity between the query and the $k$-th candidate chunk:
\[
\mathbf{h}_q,\,\mathbf{h}_k \in \mathbb{R}^d,\; s_k = \cos(\mathbf{h}_q,\mathbf{h}_k).
\]

For each $(q,\mathcal{C}(q))$, temperature scaling and negative-tier weighting are applied in logit space. Let $\tilde{s}_k$ denote the adjusted similarity:
\[
\tilde{s}_k = \begin{cases}
\; s_k/\tau, & k\in \text{P} \\
\; s_k/\tau + \log(\beta), & k\in \text{N1} \\
\; s_k/\tau + \log(\alpha), & k\in \text{N2}
\end{cases}\quad (\beta>\alpha>0).
\]
The differential weighting $\beta>\alpha$ is crucial for handling complex logical expressions, as it ensures that distractors (N1) receive stronger penalties than negatives (N2), forcing the model to learn fine-grained distinctions. The probability for a positive sample $k\in \text{P}$ is computed using only negative samples (N1 and N2) in the denominator:
\[
p_k = \frac{\exp(\tilde{s}_k)}{\sum_{t\in N1\cup N2} \exp(\tilde{s}_t) + \exp(\tilde{s}_k)}, \quad k\in \text{P}.
\]
Following the InfoNCE framework \cite{oord2018representation}, the objective minimized for a query $q$ is
\[
\mathcal{L}(q) = -\frac{1}{|P|} \sum_{k\in P} \log p_k.
\]
This design scales distractor/negative samples in logit space (with $\beta>\alpha$), excludes positives from the denominator to avoid dilution, and emphasizes the hierarchical separation $\text{P} \succ \text{N1} \succ \text{N2}$ to improve understanding of complex logical expressions.

\noindent\textbf{Training process}
The document encoder $\mathcal{M}_d$ is kept frozen with pre-trained parameters, while the query encoder $\mathcal{M}_q$ is fine-tuned during training. Based on the embeddings and similarities defined above, the contrastive loss serves as the optimization objective. Formally, the query encoder parameters $\theta_q$ are optimized by
\[
\theta_q^* = \arg\min_{\theta_q} \; \mathbb{E}_{q \sim \mathcal{D}}\, \mathcal{L}(q),
\]
where $\mathcal{D}$ denotes the training set.

\begin{figure}[htbp]
  \centering
  \includegraphics[width=0.47\textwidth]{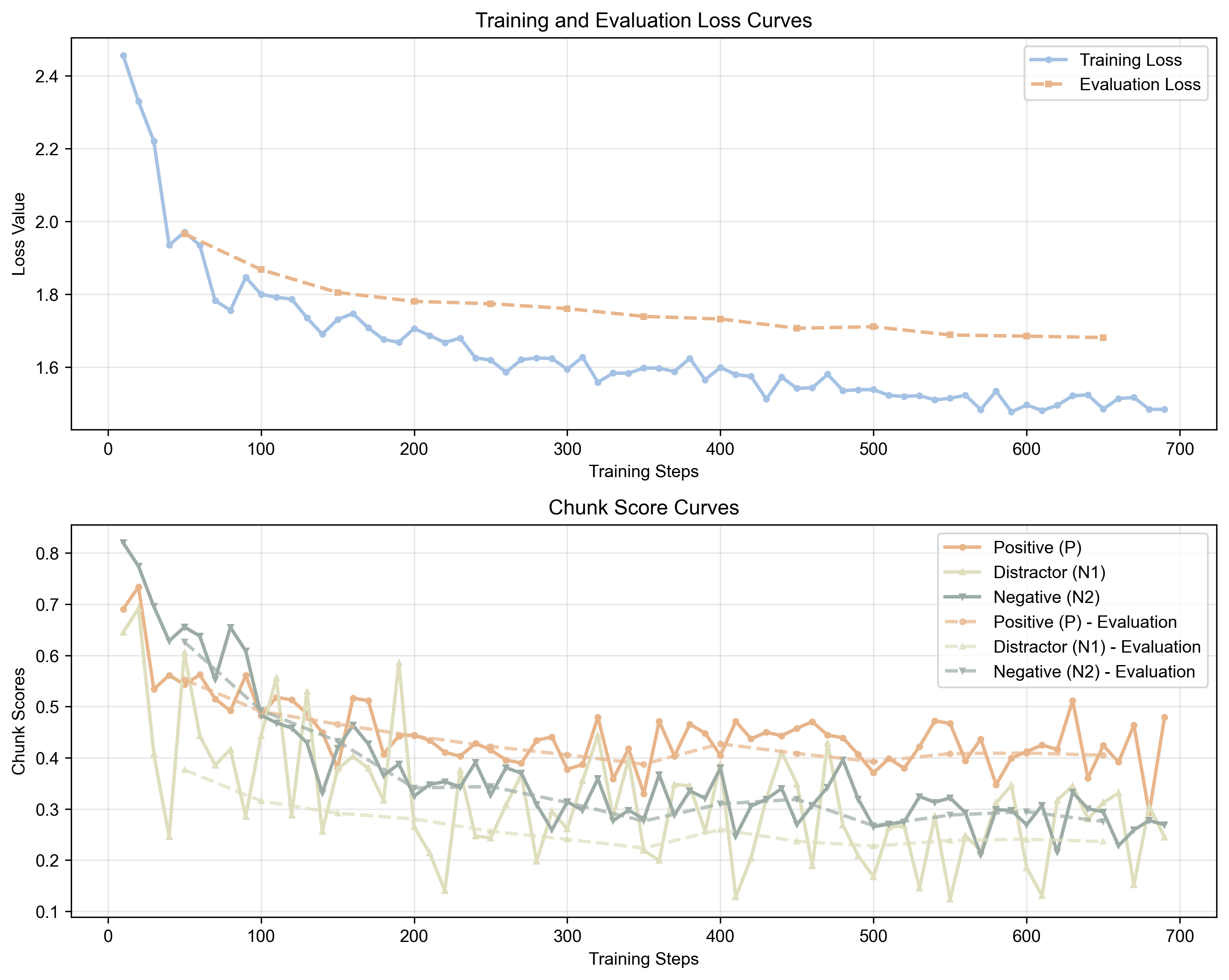}
  \caption{Training and validation loss curves (upper) and chunk score evolution for different chunk categories (lower) during Qwen3-Embedding-0.6B fine-tuning.}
  \label{fig:training_curves}
\end{figure}

\section{Experiments}

\subsection{Experimental Setup}

\noindent\textbf{Datasets}
Experiments are conducted on three widely-used public datasets: HotpotQA \cite{yang2018hotpotqa}, MS~MARCO \cite{nguyen2016msmarco}, and MuSiQue \cite{trivedi-etal-2022-musique}. For training and validation sets, Query Rewriting is performed and fine-grained category contrastive information is annotated on MS~MARCO data, while HotpotQA and MuSiQue datasets remain unannotated. For test sets, Query Rewriting is applied to all datasets while preserving the original test sets for comparison. The rewritten test sets undergo manual screening to ensure they meet the rewriting requirements.

\noindent\textbf{Models and Baselines}
Experiments are conducted using two embedding models: Qwen3 Embedding \cite{qwen3embedding} and BGE M3-Embedding \cite{bge-m3}. The effectiveness of LORE is compared against raw models and traditional contrastive learning approaches on both raw test sets and disturbed test sets.

\noindent\textbf{Evaluation Metrics}
Retrieval performance is primarily evaluated using positive recall at top-$k$ positions (@3, @5, @10), which measures the proportion of relevant chunks successfully retrieved within the top-$k$ results. For Query Rewriting test sets, the recall rates of distractor samples are also recorded.

\noindent\textbf{Experimental Setup}
Model training is conducted on a single A800 GPU. The learning rate is set to $1 \times 10^{-5}$, temperature parameter $\tau = 0.05$, and tier-weighted parameters $\alpha = 1.0$, $\beta = 3.0$. Training is performed for 1 epoch with batch size 32. For reproducibility, three models with different random seeds (0, 1, 2) are trained and the mean and standard deviation across runs are reported.

\begin{table*}[ht]
\centering
\caption{Recall of positive chunks and distractor chunks.}
\label{tab:results}
\resizebox{\textwidth}{!}{
\begin{tabular}{ccccccccccc}
\toprule
\textbf{Recall} &       & \multicolumn{3}{c}{\textbf{HotpotQA}} & \multicolumn{3}{c}{\textbf{Musique}} & \multicolumn{3}{c}{\textbf{MSMARCO}} \\
\cmidrule(lr){3-5} \cmidrule(lr){6-8} \cmidrule(lr){9-11}
     &       & \textbf{Raw Dataset}& \multicolumn{2}{c}{\textbf{Disturbed Dataset}} & \textbf{Raw Dataset}& \multicolumn{2}{c}{\textbf{Disturbed Dataset}} & \textbf{Raw Dataset}& \multicolumn{2}{c}{\textbf{Disturbed Dataset}} \\
     &       & \textbf{P$\uparrow$} & \textbf{P$\uparrow$} & \textbf{N1$\downarrow$} & \textbf{P$\uparrow$} & \textbf{P$\uparrow$} & \textbf{N1$\downarrow$} & \textbf{P$\uparrow$} & \textbf{P$\uparrow$} & \textbf{N1$\downarrow$} \\
\midrule
\multicolumn{11}{c}{\textbf{Qwen/Qwen3-Embedding-0.6B}} \\
\midrule
\textbf{@3} & \textbf{Raw Model}  & \cellcolor[HTML]{EBB87F}56.37{\scriptsize ±0.00} & \cellcolor[HTML]{B3D4DB}33.76{\scriptsize ±0.00} & \cellcolor[HTML]{B3D4DB}70.03{\scriptsize ±0.00} & \cellcolor[HTML]{EBB87F}59.07{\scriptsize ±0.00} & \cellcolor[HTML]{B3D4DB}39.55{\scriptsize ±0.00} & \cellcolor[HTML]{B3D4DB}83.88{\scriptsize ±0.00} & \cellcolor[HTML]{EBB87F}62.63{\scriptsize ±0.00} & \cellcolor[HTML]{B3D4DB}46.98{\scriptsize ±0.00} & \cellcolor[HTML]{B3D4DB}84.25{\scriptsize ±0.00} \\
& \textbf{+InfoNCE}  & \cellcolor[HTML]{EBB87F}\textbf{70.87}{\scriptsize ±0.40} & \cellcolor[HTML]{B3D4DB}59.71{\scriptsize ±0.56} & \cellcolor[HTML]{B3D4DB}50.51{\scriptsize ±1.06} & \cellcolor[HTML]{EBB87F}\textbf{67.23}{\scriptsize ±0.42} & \cellcolor[HTML]{B3D4DB}53.65{\scriptsize ±0.47} & \cellcolor[HTML]{B3D4DB}70.36{\scriptsize ±1.16} & \cellcolor[HTML]{EBB87F}\textbf{68.39}{\scriptsize ±0.25} & \cellcolor[HTML]{B3D4DB}51.63{\scriptsize ±0.40} & \cellcolor[HTML]{B3D4DB}82.03{\scriptsize ±0.56} \\
 & \textbf{+LORE} & \cellcolor[HTML]{EBB87F}70.72{\scriptsize ±0.09} & \cellcolor[HTML]{B3D4DB}\textbf{65.45}{\scriptsize ±0.54} & \cellcolor[HTML]{B3D4DB}\textbf{23.81}{\scriptsize ±1.39} & \cellcolor[HTML]{EBB87F}66.68{\scriptsize ±0.52} & \cellcolor[HTML]{B3D4DB}\textbf{59.28}{\scriptsize ±0.34} & \cellcolor[HTML]{B3D4DB}\textbf{36.87}{\scriptsize ±2.06} & \cellcolor[HTML]{EBB87F}68.35{\scriptsize ±0.23} & \cellcolor[HTML]{B3D4DB}\textbf{69.30}{\scriptsize ±0.35} & \cellcolor[HTML]{B3D4DB}\textbf{20.15}{\scriptsize ±0.87} \\
\cmidrule{1-11}
\textbf{@5} & \textbf{Raw Model}  & \cellcolor[HTML]{EBB87F}68.04{\scriptsize ±0.00} & \cellcolor[HTML]{B3D4DB}49.23{\scriptsize ±0.00} & \cellcolor[HTML]{B3D4DB}81.59{\scriptsize ±0.00} & \cellcolor[HTML]{EBB87F}70.13{\scriptsize ±0.00} & \cellcolor[HTML]{B3D4DB}56.38{\scriptsize ±0.00} & \cellcolor[HTML]{B3D4DB}88.80{\scriptsize ±0.00} & \cellcolor[HTML]{EBB87F}82.41{\scriptsize ±0.00} & \cellcolor[HTML]{B3D4DB}73.25{\scriptsize ±0.00} & \cellcolor[HTML]{B3D4DB}93.34{\scriptsize ±0.00} \\
& \textbf{+InfoNCE}  & \cellcolor[HTML]{EBB87F}\textbf{81.96}{\scriptsize ±0.20} & \cellcolor[HTML]{B3D4DB}74.77{\scriptsize ±0.68} & \cellcolor[HTML]{B3D4DB}69.65{\scriptsize ±1.13} & \cellcolor[HTML]{EBB87F}\textbf{79.93}{\scriptsize ±0.34} & \cellcolor[HTML]{B3D4DB}70.03{\scriptsize ±0.23} & \cellcolor[HTML]{B3D4DB}81.25{\scriptsize ±0.85} & \cellcolor[HTML]{EBB87F}84.96{\scriptsize ±0.03} & \cellcolor[HTML]{B3D4DB}76.68{\scriptsize ±0.24} & \cellcolor[HTML]{B3D4DB}92.71{\scriptsize ±0.05} \\
 & \textbf{+LORE} & \cellcolor[HTML]{EBB87F}81.84{\scriptsize ±0.44} & \cellcolor[HTML]{B3D4DB}\textbf{77.79}{\scriptsize ±0.34} & \cellcolor[HTML]{B3D4DB}\textbf{39.73}{\scriptsize ±2.09} & \cellcolor[HTML]{EBB87F}79.41{\scriptsize ±0.30} & \cellcolor[HTML]{B3D4DB}\textbf{73.32}{\scriptsize ±0.05} & \cellcolor[HTML]{B3D4DB}\textbf{52.81}{\scriptsize ±1.76} & \cellcolor[HTML]{EBB87F}\textbf{85.06}{\scriptsize ±0.31} & \cellcolor[HTML]{B3D4DB}\textbf{85.77}{\scriptsize ±0.22} & \cellcolor[HTML]{B3D4DB}\textbf{44.22}{\scriptsize ±0.78} \\
\cmidrule{1-11}
\textbf{@10} & \textbf{Raw Model}  & \cellcolor[HTML]{EBB87F}80.55{\scriptsize ±0.00} & \cellcolor[HTML]{B3D4DB}69.13{\scriptsize ±0.00} & \cellcolor[HTML]{B3D4DB}90.96{\scriptsize ±0.00} & \cellcolor[HTML]{EBB87F}84.28{\scriptsize ±0.00} & \cellcolor[HTML]{B3D4DB}77.46{\scriptsize ±0.00} & \cellcolor[HTML]{B3D4DB}92.09{\scriptsize ±0.00} & \cellcolor[HTML]{EBB87F}96.75{\scriptsize ±0.00} & \cellcolor[HTML]{B3D4DB}96.70{\scriptsize ±0.00} & \cellcolor[HTML]{B3D4DB}99.45{\scriptsize ±0.00} \\
& \textbf{+InfoNCE}  & \cellcolor[HTML]{EBB87F}\textbf{92.61}{\scriptsize ±0.08} & \cellcolor[HTML]{B3D4DB}89.30{\scriptsize ±0.21} & \cellcolor[HTML]{B3D4DB}86.95{\scriptsize ±0.64} & \cellcolor[HTML]{EBB87F}\textbf{92.81}{\scriptsize ±0.23} & \cellcolor[HTML]{B3D4DB}88.75{\scriptsize ±0.17} & \cellcolor[HTML]{B3D4DB}89.97{\scriptsize ±0.29} & \cellcolor[HTML]{EBB87F}96.75{\scriptsize ±0.00} & \cellcolor[HTML]{B3D4DB}96.70{\scriptsize ±0.00} & \cellcolor[HTML]{B3D4DB}99.45{\scriptsize ±0.00} \\
 & \textbf{+LORE} & \cellcolor[HTML]{EBB87F}92.45{\scriptsize ±0.02} & \cellcolor[HTML]{B3D4DB}\textbf{89.62}{\scriptsize ±0.32} & \cellcolor[HTML]{B3D4DB}\textbf{62.84}{\scriptsize ±2.28} & \cellcolor[HTML]{EBB87F}92.37{\scriptsize ±0.16} & \cellcolor[HTML]{B3D4DB}\textbf{88.87}{\scriptsize ±0.06} & \cellcolor[HTML]{B3D4DB}\textbf{74.66}{\scriptsize ±0.96} & \cellcolor[HTML]{EBB87F}96.75{\scriptsize ±0.00} & \cellcolor[HTML]{B3D4DB}96.70{\scriptsize ±0.00} & \cellcolor[HTML]{B3D4DB}99.45{\scriptsize ±0.00} \\
\midrule
\multicolumn{11}{c}{\textbf{BAAI/bge-m3}} \\
\midrule
\textbf{@3} & \textbf{Raw Model}  & \cellcolor[HTML]{EBB87F}64.68{\scriptsize ±0.00} & \cellcolor[HTML]{B3D4DB}40.04{\scriptsize ±0.00} & \cellcolor[HTML]{B3D4DB}71.84{\scriptsize ±0.00} & \cellcolor[HTML]{EBB87F}60.12{\scriptsize ±0.00} & \cellcolor[HTML]{B3D4DB}42.33{\scriptsize ±0.00} & \cellcolor[HTML]{B3D4DB}84.07{\scriptsize ±0.00} & \cellcolor[HTML]{EBB87F}68.10{\scriptsize ±0.00} & \cellcolor[HTML]{B3D4DB}48.95{\scriptsize ±0.00} & \cellcolor[HTML]{B3D4DB}84.02{\scriptsize ±0.00} \\
& \textbf{+InfoNCE}  & \cellcolor[HTML]{EBB87F}\textbf{71.53}{\scriptsize ±0.17} & \cellcolor[HTML]{B3D4DB}57.77{\scriptsize ±0.62} & \cellcolor[HTML]{B3D4DB}53.96{\scriptsize ±0.89} & \cellcolor[HTML]{EBB87F}\textbf{66.51}{\scriptsize ±0.06} & \cellcolor[HTML]{B3D4DB}52.94{\scriptsize ±0.09} & \cellcolor[HTML]{B3D4DB}73.48{\scriptsize ±0.36} & \cellcolor[HTML]{EBB87F}70.21{\scriptsize ±0.27} & \cellcolor[HTML]{B3D4DB}52.49{\scriptsize ±0.29} & \cellcolor[HTML]{B3D4DB}80.93{\scriptsize ±0.21} \\
 & \textbf{+LORE} & \cellcolor[HTML]{EBB87F}70.76{\scriptsize ±0.22} & \cellcolor[HTML]{B3D4DB}\textbf{65.57}{\scriptsize ±0.58} & \cellcolor[HTML]{B3D4DB}\textbf{20.43}{\scriptsize ±1.38} & \cellcolor[HTML]{EBB87F}65.88{\scriptsize ±0.08} & \cellcolor[HTML]{B3D4DB}\textbf{59.94}{\scriptsize ±0.53} & \cellcolor[HTML]{B3D4DB}\textbf{25.80}{\scriptsize ±2.78} & \cellcolor[HTML]{EBB87F}\textbf{70.52}{\scriptsize ±0.23} & \cellcolor[HTML]{B3D4DB}\textbf{69.28}{\scriptsize ±0.08} & \cellcolor[HTML]{B3D4DB}\textbf{22.80}{\scriptsize ±0.75} \\
\cmidrule{1-11}
\textbf{@5} & \textbf{Raw Model}  & \cellcolor[HTML]{EBB87F}76.09{\scriptsize ±0.00} & \cellcolor[HTML]{B3D4DB}58.24{\scriptsize ±0.00} & \cellcolor[HTML]{B3D4DB}82.89{\scriptsize ±0.00} & \cellcolor[HTML]{EBB87F}71.96{\scriptsize ±0.00} & \cellcolor[HTML]{B3D4DB}60.28{\scriptsize ±0.00} & \cellcolor[HTML]{B3D4DB}88.23{\scriptsize ±0.00} & \cellcolor[HTML]{EBB87F}85.25{\scriptsize ±0.00} & \cellcolor[HTML]{B3D4DB}75.20{\scriptsize ±0.00} & \cellcolor[HTML]{B3D4DB}93.51{\scriptsize ±0.00} \\
& \textbf{+InfoNCE}  & \cellcolor[HTML]{EBB87F}\textbf{83.06}{\scriptsize ±0.23} & \cellcolor[HTML]{B3D4DB}73.74{\scriptsize ±0.11} & \cellcolor[HTML]{B3D4DB}72.22{\scriptsize ±0.57} & \cellcolor[HTML]{EBB87F}\textbf{77.85}{\scriptsize ±0.05} & \cellcolor[HTML]{B3D4DB}69.19{\scriptsize ±0.16} & \cellcolor[HTML]{B3D4DB}84.02{\scriptsize ±0.20} & \cellcolor[HTML]{EBB87F}85.98{\scriptsize ±0.18} & \cellcolor[HTML]{B3D4DB}78.17{\scriptsize ±0.20} & \cellcolor[HTML]{B3D4DB}92.31{\scriptsize ±0.20} \\
 & \textbf{+LORE} & \cellcolor[HTML]{EBB87F}82.52{\scriptsize ±0.11} & \cellcolor[HTML]{B3D4DB}\textbf{77.59}{\scriptsize ±0.38} & \cellcolor[HTML]{B3D4DB}\textbf{33.05}{\scriptsize ±2.09} & \cellcolor[HTML]{EBB87F}77.28{\scriptsize ±0.23} & \cellcolor[HTML]{B3D4DB}\textbf{72.29}{\scriptsize ±0.06} & \cellcolor[HTML]{B3D4DB}\textbf{40.00}{\scriptsize ±3.06} & \cellcolor[HTML]{EBB87F}\textbf{86.20}{\scriptsize ±0.21} & \cellcolor[HTML]{B3D4DB}\textbf{86.10}{\scriptsize ±0.09} & \cellcolor[HTML]{B3D4DB}\textbf{44.89}{\scriptsize ±0.60} \\
\cmidrule{1-11}
\textbf{@10} & \textbf{Raw Model}  & \cellcolor[HTML]{EBB87F}88.28{\scriptsize ±0.00} & \cellcolor[HTML]{B3D4DB}78.86{\scriptsize ±0.00} & \cellcolor[HTML]{B3D4DB}92.47{\scriptsize ±0.00} & \cellcolor[HTML]{EBB87F}86.58{\scriptsize ±0.00} & \cellcolor[HTML]{B3D4DB}81.27{\scriptsize ±0.00} & \cellcolor[HTML]{B3D4DB}91.69{\scriptsize ±0.00} & \cellcolor[HTML]{EBB87F}96.75{\scriptsize ±0.00} & \cellcolor[HTML]{B3D4DB}96.70{\scriptsize ±0.00} & \cellcolor[HTML]{B3D4DB}99.45{\scriptsize ±0.00} \\
& \textbf{+InfoNCE}  & \cellcolor[HTML]{EBB87F}\textbf{93.26}{\scriptsize ±0.03} & \cellcolor[HTML]{B3D4DB}89.14{\scriptsize ±0.12} & \cellcolor[HTML]{B3D4DB}87.55{\scriptsize ±0.21} & \cellcolor[HTML]{EBB87F}\textbf{91.69}{\scriptsize ±0.05} & \cellcolor[HTML]{B3D4DB}87.78{\scriptsize ±0.07} & \cellcolor[HTML]{B3D4DB}90.90{\scriptsize ±0.13} & \cellcolor[HTML]{EBB87F}96.75{\scriptsize ±0.00} & \cellcolor[HTML]{B3D4DB}96.70{\scriptsize ±0.00} & \cellcolor[HTML]{B3D4DB}99.45{\scriptsize ±0.00} \\
 & \textbf{+LORE} & \cellcolor[HTML]{EBB87F}93.10{\scriptsize ±0.08} & \cellcolor[HTML]{B3D4DB}\textbf{89.49}{\scriptsize ±0.61} & \cellcolor[HTML]{B3D4DB}\textbf{54.90}{\scriptsize ±2.96} & \cellcolor[HTML]{EBB87F}91.37{\scriptsize ±0.08} & \cellcolor[HTML]{B3D4DB}\textbf{88.36}{\scriptsize ±0.09} & \cellcolor[HTML]{B3D4DB}\textbf{61.53}{\scriptsize ±2.59} & \cellcolor[HTML]{EBB87F}96.75{\scriptsize ±0.00} & \cellcolor[HTML]{B3D4DB}96.70{\scriptsize ±0.00} & \cellcolor[HTML]{B3D4DB}99.45{\scriptsize ±0.00} \\
\bottomrule
\end{tabular}
}
\end{table*}

\subsection{Results and Analysis}

\noindent\textbf{Training Dynamics} 
Figure~\ref{fig:training_curves} shows the learning progression of LORE. Initially, the model favors distractor (N1) samples over positive (P) samples due to surface-level lexical overlap, despite N1 being irrelevant for answering queries. 

During training, Distractor (N1) sample scores consistently decrease and eventually converge with negative (N2) samples at lower levels, while positive (P) samples maintain relatively stable scores and establish clear dominance. This progressive separation validates the effectiveness of the tier-weighted contrastive loss design.

\noindent\textbf{Retrieval Performance} 
As shown in Table~\ref{tab:results}, both InfoNCE and LORE significantly outperform the Raw Model on both Raw Dataset and Disturbed Dataset. Since only the query encoder $\mathcal{M}_q$ is fine-tuned in the experiments, this demonstrates that contrastive learning can enhance the model's understanding of queries, thereby improving retrieval performance. However, compared to InfoNCE, LORE, which utilizes fine-grained annotations from large language models, achieves more significant improvements.

\noindent\textbf{Generalization Capability}
Only MSMARCO received fine-grained annotations in the training set, yet both HotpotQA and MuSiQue datasets also benefit from this approach. This indicates that the LORE method possesses cross-task generalization effects. When constructing sufficiently general and rich fine-grained datasets, broader performance improvements in embedding models across various tasks can be anticipated. This transferability demonstrates the practical value of LORE in real-world applications where comprehensive annotation of all target domains is often impractical.

\section{Limitations and Conclusion}

Existing embedding models are limited by their training corpora and pretraining paradigms, resulting in insufficient comprehension of queries with complex logical expressions and an overreliance on surface-level similarity. Previous work has mainly focused on using large models in specific scenarios to enhance performance, rather than improving the foundational capabilities of embedding models. To address this, queries disturbed with complex logical expressions are constructed based on Rhetorical Structure Theory (RST), and fine-grained contrastive annotations across three chunk categories are applied. Contrastive learning enables embedding models to better capture complex logical expressions while remaining fully compatible with existing methods, reducing the need for additional task-specific designs.

Despite demonstrated effectiveness, the approach is constrained by the diversity of logical structures and datasets. Its generalization requires further investigation. Future directions include developing agent-based data annotation systems over more realistic queries with complex logical expressions, allowing the transfer of large models' query-intent understanding to smaller embedding models via contrastive learning, thereby strengthening their foundational capabilities.

\newpage
\section{Acknowledgments}
This work was supported by the Major Program of National Fund of Philosophy and Social Science of China (Grant No. 21\&ZD238).

\bibliographystyle{IEEEbib}
\bibliography{strings,refs}

\end{document}